\pdfoutput=1
\PassOptionsToPackage{table}{xcolor}

\documentclass[11pt]{article}

\usepackage[preprint]{acl}
\usepackage{times}
\usepackage{latexsym}

\usepackage[T1]{fontenc}
\usepackage[utf8]{inputenc}

\usepackage{hyperref}
\usepackage{adjustbox}
\usepackage{graphicx}

\usepackage{microtype}
\usepackage{booktabs,arydshln}
\usepackage{array}
\usepackage{ragged2e}
 \usepackage{amsmath}
 \usepackage{threeparttable}
 \usepackage{multicol}
 \usepackage{multirow}
 \usepackage{ccicons}
 \usepackage{chngcntr}

 \makeatletter
\def\adl@drawiv#1#2#3{%
        \hskip.5\tabcolsep
        \xleaders#3{#2.5\@tempdimb #1{1}#2.5\@tempdimb}%
                #2\z@ plus1fil minus1fil\relax
        \hskip.5\tabcolsep}
\newcommand{\cdashlinelr}[1]{%
  \noalign{\vskip\aboverulesep
           \global\let\@dashdrawstore\adl@draw
           \global\let\adl@draw\adl@drawiv}
  \cdashline{#1}
  \noalign{\global\let\adl@draw\@dashdrawstore
           \vskip\belowrulesep}}
\makeatother

\makeatletter
\def\adl@drawiv#1#2#3{%
  \hskip.5\tabcolsep
  {\color[gray]{0.7} 
  \xleaders#3{#2.5\@tempdimb #1{1}#2.5\@tempdimb}%
      #2\z@ plus1fil minus1fil\relax}%
  \hskip.5\tabcolsep}
\newcommand{\cdashlinelrg}[1]{%
  \noalign{\vskip\aboverulesep
           \global\let\@dashdrawstore\adl@draw
           \global\let\adl@draw\adl@drawiv}
  \cdashline{#1}
  \noalign{\global\let\adl@draw\@dashdrawstore
           \vskip\belowrulesep}}
\makeatother

\usepackage{pifont}
\usepackage{fontawesome5}

\definecolor{richgreen}{RGB}{102, 204, 102}
\definecolor{richred}{RGB}{255, 99, 71}

\definecolor{richorange}{RGB}{255, 165, 0}

\definecolor{ctxorange}{RGB}{217,95,2} 
\definecolor{ctxteal}{RGB}{27,158,119} 
\definecolor{coolgray}{RGB}{160,160,170}

\definecolor{slateblue}{RGB}{100,140,200}

\definecolor{augpurple}{RGB}{117,107,177}

\usepackage[most]{tcolorbox}   
\usepackage{siunitx}           
\usepackage{colortbl} 

\newcolumntype{R}{p{0.05cm}} 

\newcolumntype{P}[1]{>{\RaggedRight\arraybackslash}p{#1}}
\usepackage{makecell}
\newcolumntype{C}[1]{>{\centering\arraybackslash}p{#1}}
\newcolumntype{D}[1]{>{\raggedright\arraybackslash}p{#1}}

\usepackage{amsmath}
\usepackage{amsfonts}
\usepackage{inconsolata}

\usepackage{graphicx}

\usepackage{blindtext}

\usepackage{afterpage}

\usepackage[capitalize]{cleveref}
\crefname{figure}{Fig.}{Figs.}
\crefname{table}{Tab.}{Tabs.}
\crefname{appendix}{App.}{Apps.}

\usepackage{todonotes}
\usepackage{soul}

\definecolor{TodoColor}{rgb}{1,0.7,0.6}

\definecolor{OndrejColor}{rgb}{0.7,0.8,1} 

\definecolor{NickColor}{rgb}{0.7,1.0,0.7} 

\usepackage{booktabs}   
\usepackage{colortbl}   
\usepackage[table]{xcolor} 
\usepackage[table,HTML]{xcolor}

\newcommand{\parag}[1]{\vspace{1ex}\noindent\textbf{#1}}
\usepackage{enumitem}
\usepackage{placeins}
\newcommand{\huggingfacesmall}{\includegraphics[width=9px]{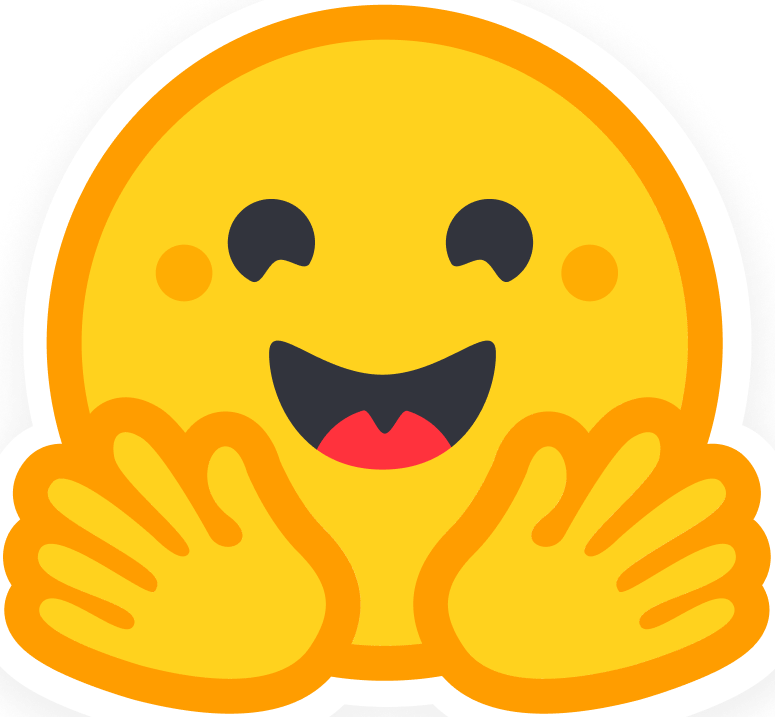}}
\newcommand{\vosksmall}{\includegraphics[width=9px]{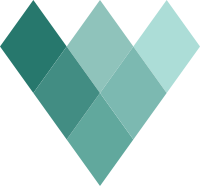}}
\newcommand{\kaldismall}{\includegraphics[width=9px]{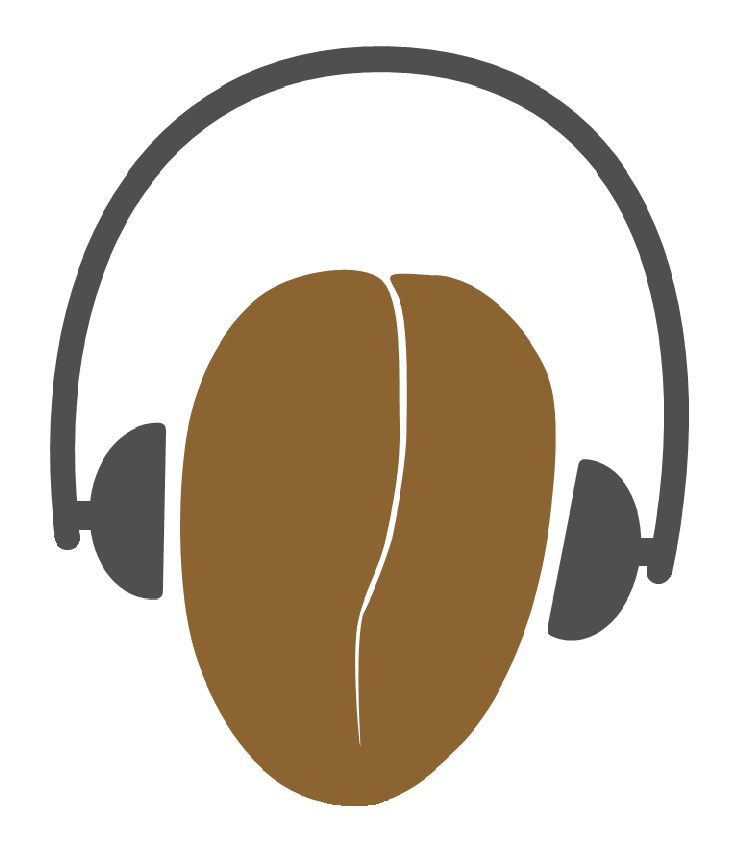}}
\newcommand{\pyannotesmall}{\includegraphics[width=9px]{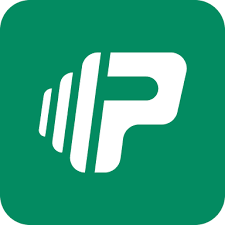}}

\usepackage{amssymb}
\usepackage{pifont}
\newcommand{\cmark}{\raisebox{0.2ex}{\scalebox{0.85}{\ding{51}}}}
\newcommand{\xmark}{\raisebox{0.2ex}{\scalebox{0.85}{\ding{55}}}}

\title{F-Actor: Controllable Conversational Behaviour in Full-Duplex Models}

\newcommand{\authorsep}{\quad}

\author{
 \textbf{Maike Züfle\textsuperscript{1}}\authorsep
 \textbf{Ondrej Klejch\textsuperscript{2}}\authorsep
 \textbf{Nicholas Sanders\textsuperscript{2}}
\\
 \textbf{Jan Niehues\textsuperscript{1}}\authorsep
 \textbf{Alexandra Birch\textsuperscript{2}}\authorsep
\textbf{Tsz Kin Lam\textsuperscript{3}\thanks{Work done while at the University of Edinburgh.}}
\\
\\
 \textsuperscript{1}Karlsruhe Institute of Technology\authorsep
 \textsuperscript{2}University of Edinburgh\authorsep
 \textsuperscript{3}NatWest
\\
 \small{\texttt{maike.zuefle@kit.edu}}
}

\begin{document}
\maketitle

\begin{abstract}
Spoken conversational systems require more than accurate speech generation to have human-like conversations: to feel natural and engaging, they must produce conversational behaviour that adapts dynamically to the context. Current spoken conversational systems, however, rarely allow such customization, limiting their naturalness and usability. In this work, we present the first open, instruction-following full-duplex conversational speech model that can be trained efficiently under typical academic resource constraints. By keeping the audio encoder frozen and finetuning only the language model, our model requires just 2,000 hours of data, without relying on large-scale pretraining or multi-stage optimization. The model can follow explicit instructions to control speaker voice, conversation topic, conversational behaviour (e.g., backchanneling and interruptions), and dialogue initiation. We propose a single-stage training protocol and systematically analyze design choices. Both the model and training code is released to enable reproducible research on controllable full-duplex speech systems.\footnote{\url{https://github.com/MaikeZuefle/f-actor}}
\end{abstract}

\section{Introduction}\label{sec:intro}
Developing a machine that can interact with humans in a natural, conversational way has been a research goal since the Dartmouth proposal in 1955 \citep{McCarthy_Minsky_Rochester_Shannon_2006}. Although today's conversational systems in text-to-text settings are approaching human-like communication 
\citep{cheng-etal-2024-anthroscore},
their speech counterparts continue to exhibit significant limitations.

\begin{figure*}
    \centering
    \includegraphics[width=1.0\linewidth]{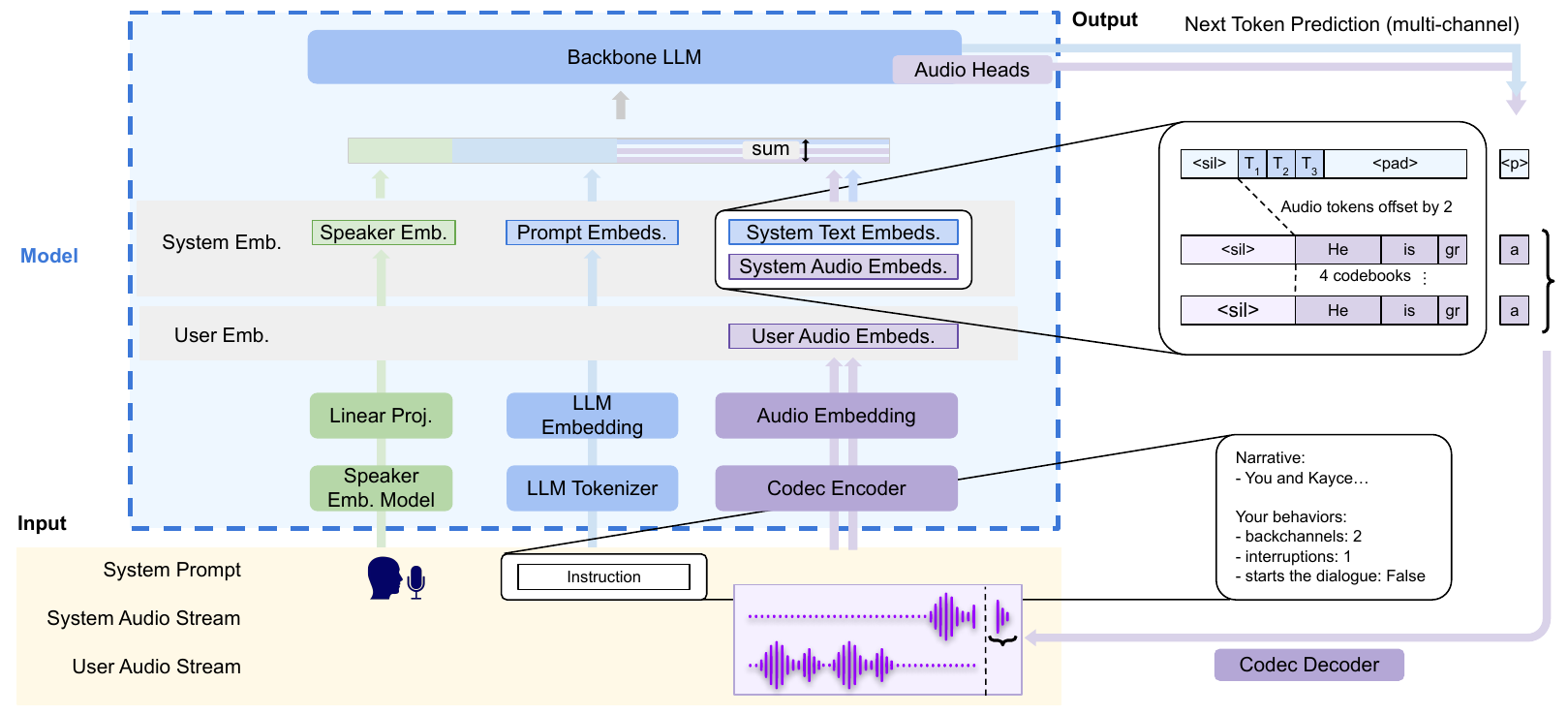}
    \caption{Overview of our controllable full-duplex model, which can be prompted to control (i) speaker voice, (ii) conversation topic, (iii) conversational behaviour (e.g., backchanneling and interruptions), and (iv) dialogue initiation. Only the LLM and audio heads are trained in a single-stage training, other components remain frozen.}
    \label{fig:architecture}
\end{figure*}

A key property of natural human conversation is its full-duplex nature \citep{Stivers2009UniversalsAC}: humans can listen and speak at the same time. 
This facilitates natural turn-taking \citep{turn_taking} by enabling interruptions and backchannels, such as brief acknowledgments or sounds of agreement produced while the other speaker continues talking.
For conversational speech systems to behave in a human-like manner, they should therefore be able to handle overlapping speech \citep{backchannel, alexa_backchannel} and be able to dynamically adapt their behaviour in real time. 

Recent work has begun to address full-duplex modelling using a variety of architectural choices: training special predictors for overlapping speech \citep{ruede2017yeahrightuhhuhdeep, chen2025minmomultimodallargelanguage}, modelling user and system speech as separate streams \citep{défossez2024moshispeechtextfoundationmodel, hu25f_interspeech} or by interleaving user and system chunks in a single sequence \citep{veluri-etal-2024-beyond, lee-etal-2025-behavior}.
While the latter approaches enable models to \emph{handle} interruptions and backchannels, they typically do not \emph{model them explicitly} on the system side. That is, models are trained to robustly handle overlapping speech produced by the user, but are rarely studied in terms of whether the system itself interrupts or backchannels. As a result, existing evaluations focus on reactive rather than proactive conversational behaviour 
\citep{peng25b_interspeech, lin2025fullduplexbenchv15evaluatingoverlap}.

In addition, most current spoken conversational models offer limited customization. Properties such as the system’s voice, conversational persona, topic framing, and interaction style, including tendencies toward backchanneling and interruption, strongly influence how appropriate and human-like a system feels in a given situation \citep{KuehneFischerZhou2021, alexa_backchannel}. Commercial systems have demonstrated the importance of such design choices, e.g. by deliberately incorporating backchannels to increase perceived naturalness \citep{google_duplex2018, Lin_2022, alexa_backchannel}. Crucially, however, the optimal conversational behaviour can vary widely across users and use cases, 
motivating the need for controllable, instruction-following full-duplex speech systems.

A small number of recent works have explored instruction-following or persona-conditioned full-duplex models, for instance by specifying a target voice \citep{chen2025minmomultimodallargelanguage, shi2025voilavoicelanguagefoundationmodels} or assigning the system a persona via prompt \citep{shi2025voilavoicelanguagefoundationmodels}. To date, these models and their code have not been publicly released, they are computationally expensive to train, and the relevant behavioural characteristics are not evaluated systematically.

In this work, we present \textit{F-Actor}, a \textbf{f}ull-duplex model that behaves like an \textbf{actor} following conversational instructions. We describe a practical approach for training it under typical academic resource constraints. By keeping the audio encoder frozen and finetuning only the LLM, our approach requires just 2,000 hours of training data and two days on four A100-40GB GPUs, without relying on large-scale pretraining or multi-stage optimization.

We demonstrate that it is possible to train a model that can follow explicit instructions regarding (i) speaker voice, (ii) conversation topic, (iii) conversational behaviour, such as backchanneling and interruptions, and (iv) dialogue initiation (user- or system-driven). \cref{fig:architecture} shows an overview of our model.
Our contributions are three-fold:
\begin{enumerate}[noitemsep]
    \item We introduce the first open, controllable full-duplex model.
    \item We systematically experiment with and analyze different design and architectural choices, such as RVQ and FSQ encoders, text alignment variants, audio delays or special tokens.
    \item We develop a single-stage training protocol yielding good models on an academic budget.
\end{enumerate}

We release both the model and the training code publicly, enabling reproducibility and further research on controllable full-duplex speech systems.

\section{Background}\label{sec:background}
This section reviews the key components of our full-duplex speech language model; alternative approaches are discussed in \cref{sec:rel_work}.

\parag{Full-Duplex Modelling.}
Full-duplex interaction, simultaneous speaking and listening, is often modelled using parallel user and system speech streams \citep{hu25f_interspeech, défossez2024moshispeechtextfoundationmodel}. To enable the backbone LLMs to process speech streams, raw audio is typically mapped to discrete acoustic units (DAUs), which can be treated as tokens in the model vocabulary. Each speech stream then emits DAUs at every timestep, including silence DAUs, and embeddings for each stream are combined by summation \citep{défossez2024moshispeechtextfoundationmodel, hu25f_interspeech}, or by token-wise fusion \citep{wang2025ntppgenerativespeechlanguage} to allow concurrent processing.
To produce speech, the generated DAUs are mapped back into speech waveforms with pre-trained vocoders.

An alternative is to interleave user and system tokens within a single stream using speaker indicators \citep{veluri-etal-2024-beyond, lee-etal-2025-behavior}. However, this approach is limited by the chunk-size for interleaving and the longer sequence length, limiting its suitability for true full-duplex scenarios such as interruptions and backchannels.

\parag{Discrete Acoustic Units.}
DAUs are typically produced by neural audio codecs such as Mimi \citep{défossez2024moshispeechtextfoundationmodel}, EnCodec \citep{defossezhigh}, or SoundStream \citep{zeghidour2021soundstream}. These codecs compress audio into discrete codes optimized for reconstruction and are widely used as both inputs and outputs in recent speech language models \citep{défossez2024moshispeechtextfoundationmodel, hu25f_interspeech}. 

The codec models represent each audio frame using multiple discrete codes, drawn from multiple codebooks. The choice of quantization determines how these codes are modelled.
Dependent codebooks, typically produced by residual vector quantization \cite[RVQ]{zeghidour2021soundstream}, require hierarchical prediction: the language model predicts the first code, while additional modules generate the remaining layers \citep{défossez2024moshispeechtextfoundationmodel, wang2023neural}. In this work, we choose independent codebooks from finite scalar quantization \cite[FSQ]{fsq_2024}, which simplify modelling by allowing all codes for a frame to be predicted jointly \citep{hu25f_interspeech}. Additional details on encoding and reconstruction are provided in \cref{app:background}.

DAUs may be integrated into an LLM either by extending its vocabulary \citep{hu25f_interspeech} or by using a dedicated embedding layer for audio tokens in parallel to the text token embedding layer. In this work, we choose the latter, which
is often more efficient, as audio codebooks are substantially smaller than text vocabularies.

\section{Related Work}\label{sec:rel_work}
Full-duplex models have gained significant attention, resulting in a diverse set of architectures.

\parag{Architectures.}
Most models rely on text-based LLM backbones, with the exception of \citet{nguyen-etal-2023-generative}.
Cascaded architectures \citep{chen2025minmomultimodallargelanguage, wang2024freezeomnismartlowlatency, chen2025fireredchatpluggablefullduplexvoice, zhang2025llmenhanceddialoguemanagementfullduplex, full_duplex_schema_2024} explicitly model turn-taking and overlapping speech. This is achieved through mechanisms such as chunk-wise state prediction \citep{wang2024freezeomnismartlowlatency}, additional prediction modules \citep{chen2025minmomultimodallargelanguage}, modified VAD \citep{chen2025fireredchatpluggablefullduplexvoice}, or control tokens \citep{zhang2025llmenhanceddialoguemanagementfullduplex, full_duplex_schema_2024, yu2024salmonnomnicodecfreellmfullduplex}.

End-to-end architectures typically support these behaviours natively and have therefore become increasingly popular. Among them, Moshi \citep{défossez2024moshispeechtextfoundationmodel} and its variants \citep{ohashi25_interspeech, shi2025voilavoicelanguagefoundationmodels} represent a common design choice: employing separate system, user, and text streams, and codec-based models for speech input and output \citep{défossez2024moshispeechtextfoundationmodel, defossezhigh, zeghidour2021soundstream}. Other approaches use continuous features at the input \citep{hu25f_interspeech, fu2025vita15gpt4olevelrealtime}, or at both input and output \citep{yu2024salmonnomnicodecfreellmfullduplex}. Some architectures avoid multiple input–output streams by interleaving modalities \citep{veluri-etal-2024-beyond, lee-etal-2025-behavior, yu2024salmonnomnicodecfreellmfullduplex}, or by embedding them jointly at the token level \citep{wang2025ntppgenerativespeechlanguage}. For more details, we refer the reader to \citet{arora2025landscapespokenlanguagemodels}.

\parag{Instruction-Following.}
While recent models emphasize different aspects of full-duplex interaction, for example, incorporating chain-of-thought reasoning \citep{arora2025chainofthoughtreasoningstreamingfullduplex}, mechanisms for handling background noise \citep{liao2025flexduopluggableenablingfullduplex}, or improved turn-taking \citep{cui2025thinktalkenhancingmeaningful}, only a few focus on instruction following. 

One such model is BeDLM \citep{lee-etal-2025-behavior}, which supports instructions with regards to the conversational narrative and speaking behaviour encoded via prompts and special tokens. However, BeDLM is primarily a dialogue generation model, as it generates both speakers rather than only the system side.
MinMo \citep{chen2025minmomultimodallargelanguage}, a cascaded speech model with duplex capabilities, can be instructed using embeddings that control emotions, dialects, speaking rate, and voice imitation. These capabilities, however, are evaluated only in MinMo’s TTS setting on in-house test sets, leaving it unclear whether they extend to the duplex model.
Finally, Voila \citep{shi2025voilavoicelanguagefoundationmodels}, an end-to-end full-duplex model, is trained for instruction following by conditioning the system on a persona via text prompts and on a specific voice via a learnable speaker embedding. However, these instruction-following abilities are not evaluated or reported.

Unfortunately, MinMo, Voila\footnote{Voila released the base model, but not the full-duplex.}, and BeDLM did not release their code or models, making direct comparisons with them impossible.

\section{Model}\label{sec:model}

\textit{F-Actor}, our full-duplex model, is based around an LLM backbone, augmented with a speaker embedding model and an audio encoder. 
Unlike previous work \citep{défossez2024moshispeechtextfoundationmodel,hu25f_interspeech}, \textit{F-Actor} is designed with instruction following in mind, that is, incorporating both a speaker embedding and an instruction-following prompt.
In the following, we describe each component of the system in detail. An overview is 
provided in \cref{fig:architecture}.

\parag{Architecture.}
We use \texttt{Llama\-3.2\--1B\--Instruct}\footnote{\huggingfacesmall{} \href{https://huggingface.co/meta-llama/Llama-3.2-1B-Instruct}{meta-llama/Llama-3.2-1B-Instruct}} \citep{grattafiori2024llama3herdmodels} as our backbone LLM.
For speech encoding and decoding, we employ \texttt{Nemo-nano-codec-22khz-0.6kbps-12.5fps}\footnote{\huggingfacesmall{} \href{https://huggingface.co/nvidia/nemo-nano-codec-22khz-0.6kbps-12.5fps}{nvidia/nemo-nano-codec-22khz\-0.6kbps-12.5fps}} \citep{nanocodec_2025}.
The codec consists of four independent codebooks of size 4032, producing four DAU streams per input audio stream, operating at a frame rate of 12.5~fps. In contrast to \citet{hu25f_interspeech}, we keep the NanoCodec encoder \emph{frozen} for efficiency and train only the LLM.

NanoCodec has two advantages: (i) it uses FSQ quantization \citep{fsq_2024}  so the four codebooks are independent, enabling parallel prediction without requiring a depth-transformer architecture \citep{défossez2024moshispeechtextfoundationmodel}, and (ii) it has shown strong performance on speech synthesis \citep{hu25f_interspeech}. 

\parag{Instruction Following Component.} 
We design our model to adhere to explicit instructions. An instruction specifies (a) the narrative of the conversation, (b) whether the system should initiate interaction, (c) the frequency of backchanneling and interruptions, and (d) the system’s voice characteristics. We encode (a)–(c) as a textual prompt embedding and (d) as a speaker embedding projected into the LLM’s token space. To obtain the speaker embeddings, we extract the first five seconds of speech from a speech sample and encode them as in \citet{shi2025voilavoicelanguagefoundationmodels} using the ECAPA-TDNN architecture \citep{dawalatabad2021ecapa}. We concatenate the speaker and textual embeddings and prepend the sequence to the audio stream during training.

For (c), we specify backchanneling and interruption frequency as exact counts rather than proportions. This choice enables controlled evaluation across full-duplex models (e.g., assessing how well a model responds when interrupted exactly 5 times and backchanneled 4 times), while remaining flexible, as counts and proportions are trivially interconvertible.

During inference, the model begins by appending its first generated token to this instruction prefix, thereby initiating the conversation according to the specified behavioural parameters.

\parag{DAU Embeddings.}
Our model processes two speech streams simultaneously: a user stream \textit{user} and a system stream \textit{sys}, enabling full-duplex modeling of overlapping speech (see \cref{sec:background}). For each stream $s \in \{\text{user}, \text{sys}\}$, NanoCodec produces four DAU sequences $\mathrm{DAU}^{s}_{1}, ..., \mathrm{DAU}^{s}_{4}$.

We embed each codebook stream $\mathrm{DAU}^{s}_{i}$ with a separate embedding layer specific to the stream (user or system) and codebook index $i$. Separate embedding matrices for the user and system streams allow the model to reliably distinguish speakers. The embedding dimension is chosen to match the token embedding dimension of the LLM backbone. Unlike \citet{hu25f_interspeech,shi2025voilavoicelanguagefoundationmodels}, who extend the LLM’s vocabulary with DAU tokens, we use dedicated embedding layers to avoid computing a large joint softmax over text tokens and DAUs, thereby significantly improving efficiency.
We then sum the codebook embeddings 
over the codebook indices and user and systems streams
to produce an input embedding vector $x$: 
\vspace{-0.3cm}
\begin{equation*}
    x = \sum_{s \in \{\text{user},\:\text{sys}\}}\sum_{i=1}^{4}\mathrm{Embed}(\mathrm{DAU}^{s}_{i})
\end{equation*}%
\vspace{-0.3cm}

Finally, we concatenate the speaker embedding, the instruction prompt, and the DAU embedding to give the input for the LLM as shown in \cref{fig:architecture}.

\parag{DAU Generation.}
On top of the LLM, we attach eight linear heads that project the last token final hidden state back into DAU logits. Let $H \in \mathbb{R}^{1 \times d}$ denote the last token's final hidden state (with hidden size $d$). Each linear head $W_{k} \in \mathbb{R}^{d \times |C|}$, $k=1,\dots,8$, maps $H$ to the logits for one DAU codebook: $\mathrm{DAU}_k = H W_k$, 
where $|C| = 4032$ is the codebook size. The eight heads correspond to the four user and four system DAU streams. During inference, only the four system heads are used to sample DAU tokens, which are decoded into a waveform via the NanoCodec decoder.

\parag{Text Stream.}
Following \citet{hu25f_interspeech,défossez2024moshispeechtextfoundationmodel}, we not only generate audio, but also experiment with generating the corresponding text, using an additional system-side text stream. To do so, we need to align the text tokens and audio DAUs (alignment strategies are described in \cref{sec:experiments}). After embedding the text stream, we add it to the system’s speech stream. The LLM  uses this summed embedding as input. We then use the LLM’s original language-modeling head to predict the next text token.

\section{Evaluation}\label{sec:eval}
We evaluate our full-duplex models along two criteria: general system capabilities and instruction-following performance. Implementation details of the metrics can be found in \cref{app:eval}.

\subsection{Dialogue Generation for Evaluation.}
To evaluate these capabilities, we generate dialogues between two instances of our model, allowing them to \textit{talk to each other} \citep{veluri-etal-2024-beyond}, assigning each model instance its respective personalized instructions.

\subsection{General System Capabilities Eval.}
Following prior full-duplex work \citep{arora2025chainofthoughtreasoningstreamingfullduplex, zhang-etal-2025-omniflatten}, we report perplexity on the speech and text streams. We assess speech quality with UTMOS \citep{utmos_2022}, a neural, reference-free metric that predicts mean opinion scores to estimate perceived speech naturalness and quality. To evaluate dialogue behaviour, we measure speaking-time balance between speakers and compute WER between transcribed generated audio and generated text to assess how understandable the audio is and how coherent the text and audio streams are. Note, that the model is conditioned only on the dialogue topic and is free to generate any plausible conversation for that topic, rather than reproducing the test set dialogues. Consequently, WER is computed only between the generated text and the transcription of the generated audio, and not against the test set transcripts.

\subsection{Instruction-Following Capabilities Eval.}
We evaluate instruction-following performance along four dimensions:  
(1) \textit{Speaker Initiation}: Accuracy in initiating the conversation according to the  prompt.
(2) \textit{Speaker Embedding Consistency}: Cosine similarity between the target speaker embedding and the generated speech, averaged across speakers and conversations. To detect potential drift, we also compute the distance (1–cosine similarity) between the first and last segments for each speaker.
(3) \textit{Narrative Adherence}: An LLM judge (\texttt{Llama-3.1-8B-Instruct}\footnote{\huggingfacesmall{} \href{https://huggingface.co/meta-llama/Llama-3.1-8B-Instruct}{meta-llama/Llama-3.1-8B-Instruct}} \citep{grattafiori2024llama3herdmodels}
) evaluates the alignment (of the transcript) with the narrative specified in the prompt (prompts in \cref{fig:app_llm_judge}). We run the judge with three different seeds and report the mean and standard deviation across these. We also run a human evaluation to confirm the reliability of the judge. Details can be found in \cref{subsec:if_results}.
(4) \textit{Backchannels (BCs) and Interruptions}: We measure the correlation between prompt-specified and generated counts using Pearson’s r, with two-sided p-values computed from the exact distribution.

For (4), we experiment with different algorithms for detecting BCs and interruptions. We initially adopted the FD-Bench \citep{peng25b_interspeech} implementation, which uses Silero-VAD \citep{Silero_VAD}, but default thresholds did not generalize well to the dataset used in this work \citep[details in \cref{sec:experiments}]{lee-etal-2025-behavior}, often misclassifying interruptions or BCs. Detecting these events is challenging because short BCs and overlapping speech can be easily missed or misaligned with transcripts. We compare timestamps from Silero-VAD, Parakeet\footnote{\huggingfacesmall{} \href{https://huggingface.co/nvidia/parakeet-tdt-0.6b-v2}{nvidia/parakeet-tdt-0.6b-v2}}, and Vosk\footnote{\vosksmall \href{https://alphacephei.com/vosk/models/vosk-model-en-us-0.22.zip}{vosk-model-en-us-0.22}}, performing a grid search over interruption and overlap threshold parameters. Parakeet achieves the most reliable performance and is used for all subsequent evaluations. Detailed results for all alignment methods are reported in \cref{tab:bc_inter_algo}, with grid-search details provided in \cref{app:bc_expl}.

\section{Experimental Setup}\label{sec:experiments}
In this section, we describe the training data, setup and evaluation protocol of our full-duplex model.

\subsection{Data}

We use the Behavior-SD dataset \citep{lee-etal-2025-behavior}, which contains 2,164 hours of English multi-turn, two-speaker conversations with annotations for narrative structure, backchannels (BCs), and interruptions. Behavior-SD is generated using CosyVoice TTS \citep{du2024cosyvoicescalablemultilingualzeroshot} and, despite being synthetic, the authors claim that it exhibits strong emotion appropriateness and dialogue naturalness in human evaluations \citep{lee-etal-2025-behavior}. While real conversational speech would be preferable, large-scale, wideband, full-duplex datasets do not exist \citep{chen2025turntakingsynchronousdialoguesurvey}, especially with BC and interruption annotations, making synthetic data a practical choice. 
We extend the data in the following ways to serve our needs for instruction-following models.

\parag{Text-Speech Alignment.}
As described in \cref{sec:model}, we experiment with generating not only speech, but also the corresponding text. We explore both word- and utterance-level alignment of speech and text, as prior work has shown mixed results \citep{défossez2024moshispeechtextfoundationmodel, hu25f_interspeech}. To this end, we align text and speech tokens using forced alignment with Kaldi \citep{povey2011kaldi} and a public model\footnote{\kaldismall\phantom{ }\href{https://kaldi-asr.org/models/m13}{kaldi-model-m13}} trained on Librispeech~\cite{panayotov2015librispeech}. We exclude conversations with text–speech mismatches due to occasional TTS deviations from the transcript, resulting in 74.4\% of the original training set.

\begin{table*}[ht]
    \centering
    \footnotesize
    \begin{tabular}{cccccccccccccc}
        \toprule

        \multirow{2}{*}{\textbf{Loss}} &
        \textbf{Spk.} &
        \textbf{Text} &
        \textbf{Text} &
        \textbf{Audio} &
        \textbf{PPL} &
        \textbf{PPL} &
        \textbf{UTMOS} &
        \textbf{WER \%} &
        \textbf{Avg. Speaking} \\
                &
                \textbf{Emb.} & 
        \textbf{Stream}&

        \textbf{Align.} &
        \textbf{Delay} &
        \textbf{DAU $\downarrow$} &
        \textbf{Text $\downarrow$} &
        $\uparrow$ &
        \textbf{audio/text $\downarrow$} &
        \textbf{Diff (s) $\downarrow$} \\
         \midrule
         \multicolumn{5}{c}{Behaviour-SD Dialogues Testset} & - & -  & 3.78 & 4.5 & 15.44 \\
         \midrule
s/u & \cmark & \xmark & n/a & n/a & \cellcolor[HTML]{fafbfe}\textcolor[HTML]{000000}{25.23} & n/a & \cellcolor[HTML]{f6f9fe}\textcolor[HTML]{000000}{3.20} & n/a & \cellcolor[HTML]{a4c2f4}\textcolor[HTML]{000000}{\textbf{10.69}} \\
s/u & \cmark & \cmark & utt. & 0 & \cellcolor[HTML]{c4d7f8}\textcolor[HTML]{000000}{22.85} & \cellcolor[HTML]{dde8fb}\textcolor[HTML]{000000}{1.48} & \cellcolor[HTML]{cedef9}\textcolor[HTML]{000000}{3.33} & \cellcolor[HTML]{dce7fb}\textcolor[HTML]{000000}{23.19} & \cellcolor[HTML]{b6cef6}\textcolor[HTML]{000000}{11.50} \\
s/u & \cmark & \cmark & utt. & 1 & \cellcolor[HTML]{c4d7f8}\textcolor[HTML]{000000}{22.85} & \cellcolor[HTML]{e8f0fc}\textcolor[HTML]{000000}{1.51} & \cellcolor[HTML]{bcd2f7}\textcolor[HTML]{000000}{3.39} & \cellcolor[HTML]{c7daf8}\textcolor[HTML]{000000}{17.35} & \cellcolor[HTML]{c7d9f8}\textcolor[HTML]{000000}{12.25} \\
s/u & \cmark & \cmark & utt. & 2 & \cellcolor[HTML]{bad1f7}\textcolor[HTML]{000000}{22.43} & \cellcolor[HTML]{e8f0fc}\textcolor[HTML]{000000}{1.51} & \cellcolor[HTML]{b6cef6}\textcolor[HTML]{000000}{3.41} & \cellcolor[HTML]{c0d5f7}\textcolor[HTML]{000000}{15.18} & \cellcolor[HTML]{b2cbf6}\textcolor[HTML]{000000}{11.30} \\
s/u & \cmark & \cmark & word & 0 & \cellcolor[HTML]{ffffff}\textcolor[HTML]{000000}{25.47} & \cellcolor[HTML]{a4c2f4}\textcolor[HTML]{000000}{\textbf{1.33}} & \cellcolor[HTML]{ffffff}\textcolor[HTML]{000000}{3.17} & \cellcolor[HTML]{ffffff}\textcolor[HTML]{000000}{33.40} & \cellcolor[HTML]{c6d9f8}\textcolor[HTML]{000000}{12.21} \\
s/u & \cmark & \cmark & word & 1 & \cellcolor[HTML]{eef4fd}\textcolor[HTML]{000000}{24.71} & \cellcolor[HTML]{d9e6fa}\textcolor[HTML]{000000}{1.47} & \cellcolor[HTML]{d5e3fa}\textcolor[HTML]{000000}{3.31} & \cellcolor[HTML]{cfdff9}\textcolor[HTML]{000000}{19.53} & \cellcolor[HTML]{eef4fd}\textcolor[HTML]{000000}{14.00} \\
s/u & \cmark & \cmark & word & 2 & \cellcolor[HTML]{ccddf9}\textcolor[HTML]{000000}{23.22} & \cellcolor[HTML]{ffffff}\textcolor[HTML]{000000}{1.57} & \cellcolor[HTML]{b9d0f7}\textcolor[HTML]{000000}{3.40} & \cellcolor[HTML]{a8c5f4}\textcolor[HTML]{000000}{\phantom{1}8.35} & \cellcolor[HTML]{edf3fd}\textcolor[HTML]{000000}{13.93} \\
s/u & \xmark & \cmark & word & 2 & \cellcolor[HTML]{e1ebfb}\textcolor[HTML]{000000}{24.14} & \cellcolor[HTML]{ffffff}\textcolor[HTML]{000000}{1.57} & \cellcolor[HTML]{a4c2f4}\textcolor[HTML]{000000}{\textbf{3.47}} & \cellcolor[HTML]{a4c2f4}\textcolor[HTML]{000000}{\phantom{1}\textbf{7.16}} & \cellcolor[HTML]{e9f0fc}\textcolor[HTML]{000000}{13.77} \\
s & \cmark & \cmark & word & 2 & \cellcolor[HTML]{a4c2f4}\textcolor[HTML]{000000}{\textbf{21.45}} & \cellcolor[HTML]{f8fafe}\textcolor[HTML]{000000}{1.55} & \cellcolor[HTML]{b6cef6}\textcolor[HTML]{000000}{3.41} & \cellcolor[HTML]{a5c3f4}\textcolor[HTML]{000000}{\phantom{1}7.59} & \cellcolor[HTML]{e1ebfb}\textcolor[HTML]{000000}{13.42} \\
s/u & \cmark & \cmark (BC/I tok.)  & word & 2 & \cellcolor[HTML]{d0dff9}\textcolor[HTML]{000000}{23.38} & \cellcolor[HTML]{fbfdff}\textcolor[HTML]{000000}{1.56} & \cellcolor[HTML]{bcd2f7}\textcolor[HTML]{000000}{3.39} & \cellcolor[HTML]{abc7f5}\textcolor[HTML]{000000}{\phantom{1}9.19} & \cellcolor[HTML]{ffffff}\textcolor[HTML]{000000}{14.75} \\
         \bottomrule
    \end{tabular}
    \caption{General modeling results for full-duplex models trained to predict both system and user (s/u) or only the system role (s). Models are compared with and without a text stream, using word- or utterance-level alignment, and with different audio delays relative to the text. \textit{BC/I tok} indicate special tokens for backchannels and interruptions.}
    \label{tab:overall_results}
\end{table*}

\parag{Instruction Following Prefix.}\label{subsec:prompts}
To train the model as an instruction-following full-duplex system, we construct a prompt, similar to \citet{lee-etal-2025-behavior}, specifying (i) the number of system interruptions and BCs in the conversation, provided by Behavior-SD; (ii) whether the system should initiate the dialogue or wait for the user; and (iii) the narrative of the conversation. For the latter, we use the annotated narrative in Behavior-SD, and transform it into a narrative from the system's perspective to eliminate data bias: In roughly 85\% of the cases the narrative begins with the speaker who starts the conversation, which biases the model to rely on the narrative rather than instruction (ii) to decide on conversation initiation.
Rewriting this narrative was performed using \texttt{Gemma-1.1-7b-it}\footnote{\huggingfacesmall{} \href{https://huggingface.co/google/gemma-1.1-7b-it}{google/gemma-1.1-7b-it}} \citep{gemmateam2024gemmaopenmodelsbased} for training, and GPT-5.1 for the test set. 
The latter was chosen to provide higher-quality data for evaluation.
Prompts and example narratives are provided in \cref{fig:app_rewriting_prompts,fig:app_prompts}.

\parag{Speaker Embedding.} 
We condition the system on speaker embeddings extracted from the relatively small set of Behavior-SD speakers (52 speakers). By restricting training and evaluation to this closed speaker set, we prevent our model being used for voice cloning or impersonation. Details on the extraction and its integration are in \cref{sec:model}.

\subsection{Training Details}

To maximize efficiency, our approach utilizes a single training stage that avoids additional speech or text pretraining as required by prior models such as \citet{défossez2024moshispeechtextfoundationmodel}. 
Only the LLM backbone and the linear DAU heads are trained, while the speech encoder remains frozen. Training is performed on four NVIDIA A100-SXM4-40GB GPUs with early stopping based on validation performance, resulting in around 48 hours of training per model. We use a maximum sequence length of 2{,}048 tokens, a per-GPU batch size of 1, and 8 gradient accumulation steps. Training and inference parameters are provided in \cref{tab:hyperparams} in \cref{app:training}.

\subsection{Experiments}
We experiment with a range of architectural and training choices for our full-duplex model.

\parag{Loss on system (s) and user (u).} We train models that either predict only the system role (s) or both user and system roles (s/u). Accordingly, the training loss is computed either only on the system tokens (s) or on both user and system tokens (s/u).

\parag{Text-Stream.}
We examine the effect of adding a text stream on the system side (see \cref{sec:model}), enabling the model to predict not only audio but also the corresponding text \citep{défossez2024moshispeechtextfoundationmodel, hu25f_interspeech}. We further experiment with introducing special tokens that precede BCs  and interruptions (BC/I tok.) to help the model generate the numbers specified in the prompt.

\parag{Text Alignment.}
We experiment with different alignment strategies for the text stream, using either utterance-level (padding after each utterance) or word-level alignment (padding after each word).

\parag{Audio Delay.}
We experiment with delaying the audio stream by up to two tokens relative to the text stream, generating text tokens first while padding the audio stream before audio token generation begins \citep{défossez2024moshispeechtextfoundationmodel, hu25f_interspeech}.

\parag{RVQ vs.\ FSQ.}
All primary experiments use an FSQ encoder to benefit from independent codebooks. To test the generality of our multi-head DAU prediction architecture, we also evaluate an RVQ-based setup using \texttt{Mimi}\footnote{\huggingfacesmall{} \href{https://huggingface.co/kyutai/mimi}{kyutai/mimi}} \citep{défossez2024moshispeechtextfoundationmodel} with 8 codebooks. This allows us to verify whether predicting DAUs for each codebook in parallel remains effective with RVQ quantization.

\subsection{Baselines}
The only available instruction-following models, BeDLM, MinMo, and Voila (\cref{sec:rel_work}), do not release code or model checkpoints, making direct comparison infeasible. For instruction-following, we therefore use the Behavior-SD test dialogues as a \textit{topline} across all metrics and report metric-specific \textit{bottomlines}.

For turn-taking behaviour, we compare our model against Moshi \citep{defossezhigh} and dGLSM \citep{nguyen-etal-2023-generative}, reporting Inter-Pausal Units (IPUs), intra-turn pauses, between-turn gaps, and overlaps per minute following their reports and definitions and VAD implementation provided by \citet{nguyen-etal-2023-generative}, using Pyannote\footnote{\pyannotesmall\phantom{ }\href{https://github.com/pyannote/pyannote-audio}{pyannote-audio}} \citep{Plaquet23, Bredin23}.

\section{Results}\label{sec:results}
This section discusses the results of our full-duplex models in terms of general modeling abilities, instruction following, comparison with other SOTA models and the influence of sampling parameters. 

\subsection{General Modeling Abilities}
All results for general modeling abilities are shown in \cref{tab:overall_results}. Introducing a text stream alongside audio generation substantially improves performance across all metrics, reducing perplexity from $25.23$ to $21$–$23$. Word-level alignment between audio and text further improves synchronization compared to utterance-level, reducing WER between generated text and transcribed audio from over $23$\% to approximately $7.2$\% when combined with an audio delay of $2$ tokens. This delay consistently improves WER across settings without degrading speech naturalness, with UTMOS remaining stable around $3.4$–$3.5$. Omitting the speaker embedding slightly improves speech quality (UTMOS = $3.47$). 

 \begin{table*}[ht]
    \centering
    \resizebox{\linewidth}{!}{%
    \begin{tabular}{ccccccccccccc}
        \toprule

        \multirow{2}{*}{\textbf{Loss}} &
       \textbf{Spk.} &
        \textbf{Text} &
        \textbf{Text} &
        \textbf{Audio} &
        \textbf{Correct Start} & \textbf{Spk. Sim.} & \textbf{Spk. Drift} &
         \textbf{Narrative} &
        \textbf{BC Corr.*} & \textbf{Inter. Corr.*} \\

        &
        \textbf{Emb.} & 
          \textbf{Stream}&

        \textbf{Align.} &
                
        \textbf{Delay} &
        \textbf{(\% $\uparrow$)} & \textbf{(cos $\uparrow$)} & \textbf{(1-cos $\downarrow$)} &
        \textbf{(1--5 $\uparrow$)}  &
        \textbf{(per dial. $\uparrow$)} & \textbf{(per dial.$\uparrow$)} \\
         \midrule
          \multicolumn{5}{c}{Behaviour-SD Dialogues Testset} & 100.00 & 0.62 & 0.62 & 3.90$_{\pm0.03}$ & 0.92 & 0.74 \\
         \multicolumn{5}{c}{Lower Baseline} & 50.0$^a$ & 0.35$^b$ & 0.75$^c$ &  1.26$_{\pm0.02}$$^d$ & - & -\\
         \midrule

s/u & \cmark & \xmark & nan & - & \cellcolor[HTML]{ffffff}\textcolor[HTML]{000000}{54.0} & \cellcolor[HTML]{aec9f5}\textcolor[HTML]{000000}{0.52} & \cellcolor[HTML]{f0f5fd}\textcolor[HTML]{000000}{0.66} & \cellcolor[HTML]{ffffff}\textcolor[HTML]{000000}{1.50$_{\pm0.03}$} & \cellcolor[HTML]{a7c4f4}\textcolor[HTML]{000000}{0.56} & \cellcolor[HTML]{cedef9}\textcolor[HTML]{000000}{0.21} \\
s/u &\cmark&\cmark & utt. & 0 & \cellcolor[HTML]{bad1f7}\textcolor[HTML]{000000}{88.76} & \cellcolor[HTML]{aec9f5}\textcolor[HTML]{000000}{0.52} & \cellcolor[HTML]{ffffff}\textcolor[HTML]{000000}{0.67} & \cellcolor[HTML]{d3e2fa}\textcolor[HTML]{000000}{2.12$_{\pm0.01}$} & \cellcolor[HTML]{ffffff}\textcolor[HTML]{000000}{0.31} & \cellcolor[HTML]{ffffff}\textcolor[HTML]{000000}{0.13} \\
s/u &\cmark&\cmark & utt. & 1 & \cellcolor[HTML]{a6c3f4}\textcolor[HTML]{000000}{99.0} & \cellcolor[HTML]{a4c2f4}\textcolor[HTML]{000000}{\textbf{0.54}} & \cellcolor[HTML]{c2d6f8}\textcolor[HTML]{000000}{0.63} & \cellcolor[HTML]{cbdcf9}\textcolor[HTML]{000000}{2.24$_{\pm0.03}$} & \cellcolor[HTML]{e0eafb}\textcolor[HTML]{000000}{0.4} & \cellcolor[HTML]{c8daf8}\textcolor[HTML]{000000}{0.22} \\
s/u &\cmark&\cmark & utt. & 2 & \cellcolor[HTML]{a5c2f4}\textcolor[HTML]{000000}{99.6} & \cellcolor[HTML]{a9c5f5}\textcolor[HTML]{000000}{0.53} & \cellcolor[HTML]{a4c2f4}\textcolor[HTML]{000000}{\textbf{0.61}} & \cellcolor[HTML]{c6d9f8}\textcolor[HTML]{000000}{2.30$_{\pm0.00}$} & \cellcolor[HTML]{eef3fd}\textcolor[HTML]{000000}{0.36} & \cellcolor[HTML]{edf3fd}\textcolor[HTML]{000000}{0.16} \\
s/u &\cmark&\cmark & word & 0 & \cellcolor[HTML]{b8cff6}\textcolor[HTML]{000000}{89.8} & \cellcolor[HTML]{aec9f5}\textcolor[HTML]{000000}{0.52} & \cellcolor[HTML]{f0f5fd}\textcolor[HTML]{000000}{0.66} & \cellcolor[HTML]{eef4fd}\textcolor[HTML]{000000}{1.74$_{\pm0.02}$} & \cellcolor[HTML]{b2cbf6}\textcolor[HTML]{000000}{0.53} & \cellcolor[HTML]{c2d6f8}\textcolor[HTML]{000000}{0.23} \\
s/u &\cmark&\cmark & word & 1 & \cellcolor[HTML]{a5c2f4}\textcolor[HTML]{000000}{99.6} & \cellcolor[HTML]{a4c2f4}\textcolor[HTML]{000000}{\textbf{0.54}} & \cellcolor[HTML]{c2d6f8}\textcolor[HTML]{000000}{0.63} & \cellcolor[HTML]{d3e2fa}\textcolor[HTML]{000000}{2.12$_{\pm0.03}$} & \cellcolor[HTML]{aec9f5}\textcolor[HTML]{000000}{0.54} & \cellcolor[HTML]{a4c2f4}\textcolor[HTML]{000000}{\textbf{0.28}} \\
s/u &\cmark&\cmark & word & 2 & \cellcolor[HTML]{a4c2f4}\textcolor[HTML]{000000}{99.8} & \cellcolor[HTML]{a4c2f4}\textcolor[HTML]{000000}{\textbf{0.54}} & \cellcolor[HTML]{a4c2f4}\textcolor[HTML]{000000}{\textbf{0.61}} & \cellcolor[HTML]{a4c2f4}\textcolor[HTML]{000000}{\textbf{2.78$_{\pm0.02}$}} & \cellcolor[HTML]{aec9f5}\textcolor[HTML]{000000}{0.54} & \cellcolor[HTML]{b6cef6}\textcolor[HTML]{000000}{0.25} \\
s/u  &\xmark&\cmark & word & 2 & \cellcolor[HTML]{a5c3f4}\textcolor[HTML]{000000}{99.4} & \cellcolor[HTML]{ffffff}\textcolor[HTML]{000000}{0.36} & \cellcolor[HTML]{d1e0f9}\textcolor[HTML]{000000}{0.64} & \cellcolor[HTML]{b6cef6}\textcolor[HTML]{000000}{2.53$_{\pm0.05}$} & \cellcolor[HTML]{bcd2f7}\textcolor[HTML]{000000}{0.5} & \cellcolor[HTML]{dbe7fb}\textcolor[HTML]{000000}{0.19} \\
s &\cmark&\cmark & word & 2 & \cellcolor[HTML]{a5c2f4}\textcolor[HTML]{000000}{99.6} & \cellcolor[HTML]{a4c2f4}\textcolor[HTML]{000000}{\textbf{0.54}} & \cellcolor[HTML]{d1e0f9}\textcolor[HTML]{000000}{0.64} & \cellcolor[HTML]{bbd2f7}\textcolor[HTML]{000000}{2.46$_{\pm0.06}$} & \cellcolor[HTML]{a4c2f4}\textcolor[HTML]{000000}{\textbf{0.57}} & \cellcolor[HTML]{e7effc}\textcolor[HTML]{000000}{0.17} \\
s/u &\cmark&\cmark (BC/I tok.)  & word & 2 & \cellcolor[HTML]{a4c2f4}\textcolor[HTML]{000000}{\textbf{100.0}} & \cellcolor[HTML]{a4c2f4}\textcolor[HTML]{000000}{\textbf{0.54}} & \cellcolor[HTML]{b3ccf6}\textcolor[HTML]{000000}{0.62} & \cellcolor[HTML]{b9d0f7}\textcolor[HTML]{000000}{2.49$_{\pm0.02}$} & \cellcolor[HTML]{c3d7f8}\textcolor[HTML]{000000}{0.48} & \cellcolor[HTML]{a4c2f4}\textcolor[HTML]{000000}{\textbf{0.28}} \\

         \bottomrule
         \multicolumn{11}{l}{
        $^{a}$ random spk. starts; 
        $^{b}$ swapped spk. embeddings; 
        $^{c}$ first embedding = spk. 1, last = spk. 2; 
        $^{d}$ random narrative.} \\
         \multicolumn{11}{l}{
        $*$ All correlations are statistically significant with $p<0.01$}\\
    \end{tabular}%
    }\vspace{-0.1cm}
    \caption{Instruction-following results for full-duplex models trained to predict both system and user (s/u) or only the system role (s). Models are compared with and without a text stream, using word- or utterance-level alignment, and with different audio delays relative to the text. \textit{BC/I tok} indicate special tokens for backchannels and interruptions. The narrative was judge by an LLM judge (details in \cref{sec:eval}) with three different seeds, we report the mean and standard deviation across these.}\vspace{-0.3cm}
    \label{tab:if_results}
\end{table*}
 
The best overall trade-off is achieved using word-level alignment and audio delay of $2$ under the system-only loss, yielding the lowest perplexity ($21.45$) and near-minimal WER ($7.59$\%) while maintaining competitive speech quality and balanced speaking time of both speakers. We observe that at least 1\% of WER arises from words seen fewer than 100 times in the training data; although generated correctly in the text stream, they are mispronounced and therefore mistranscribed.

For reference,  Behaviour-SD Dialogues represent ground-truth speech, achieving UTMOS of $3.78$ and WER of $4.5$\%. While our full-duplex models do not match these oracle values, the results demonstrate the effectiveness of the proposed training strategy.

\subsection{Instruction Following}\label{subsec:if_results}
\cref{tab:if_results} summarizes our models’ performance across instruction-following metrics. We find that generally, the model trained on predicting both the system and user roles, with a text-stream, special interruption and backchannel tokens, and an audio delay of 2 performs the best.

\parag{Similarity to Prompted Speaker.}
Incorporating speaker embeddings produces speech that matches the target speaker with up to $54$ points cosine similarity, independent of other configurations. Likewise, speaker drift is minimal, in all configurations equal to or lower ($0.61$) than in the Behavior-SD test set ($0.62$), indicating a stable speaker voice. 

\parag{Correct Start.} 
The ability to accurately predict whether the model should initiate the conversation strongly depends on whether the model also generates text. Such models, in combination with audio delay, achieve over $99$\% accuracy, whereas models generating only DAUs perform similar to random.
Based on manual inspection, we find that the model follows the prompt: if both speakers are set to start, speech overlaps, whereas if no one is set to start, a 1-second pause precedes the first speaker.

\parag{Dialogue Narrative Adherence.} 
In terms of following the narrative specified in the prompt, we observe a clear benefit from incorporating a text stream and audio delay. Both word- and utterance-level alignments perform comparably, with slightly improved adherence for longer audio delays. Restricting training to the system role alone results in a small degradation in narrative adherence, likely because the model does not explicitly represent the interlocutor’s turns, making it more difficult to sustain a coherent, narrative-aligned dialogue.

\begin{table}[h]
    \centering
    \resizebox{\linewidth}{!}{%
    \begin{tabular}{lcc}
    \toprule
         \textbf{Model} & \textbf{Human Rating} & \textbf{LLM Judge}  \\
    \midrule
    Behaviour-SD Dialogues     &  4.17 & 3.88\\
    Model 2 audio delay & 3.00 & 2.68 \\
    Model 2 audio delay + BC/I tok. & 2.68 & 2.33 \\
    \bottomrule
    \end{tabular}
    }
    \caption{Human Evaluation to rate the narrative of the generated dialogues on 80 dialogues. The human rating confirms the ranking of the LLM judge.}
    \label{tab:human_eval}
\end{table}

To confirm the reliability of the LLM judge, we conduct a human evaluation to analyse its performance. We select our two best-performing models (both using s/u loss, speaker embeddings, word alignment, and an audio delay of 2; one also incorporating BC and interruption tokens) alongside the Behaviour-SD original dialogues.  Using the Pearmut platform \citep{zouhar2026pearmuthumanevaluationtranslation}, 12 annotators each rated 60 samples (20 dialogues × 3 models), with every dialogue annotated by 3 annotators, yielding 240 annotated instances across 80 unique dialogues. \cref{tab:human_eval} shows the results of this evaluation. Specifically, the group-by-system correlation between the LLM judge and human annotators is $\tau=1.00$, confirming that the LLM ranking of the three systems perfectly matches the human ranking. At the item level, the group-by-item correlation is $\tau=0.507$, indicating moderate but solid agreement on individual dialogues. Similarly, human annotators showed substantial agreement with each other, with a mean Kendall's $\tau$ of 0.517 at the item level (across 240 reviewer pairs) and a  $\tau=1.00$ at the system level (across 3 reviewer pairs), meaning all annotators ranked the three systems in the same order. More details on the human evaluation can be found in \cref{app:human_eval}.

\parag{Backchannels and Interruptions.}
Backchannels (BC) show a clear correlation between the frequency specified in the prompt and those generated by the model, particularly for word-level alignments. The overall best configuration with 2 token audio delay, yields a correlation of $0.54$. Correlation for interruptions is lower ($0.25$). This is likely because interruptions are rare, averaging only 0.9 per conversation in the training set. Having special tokens preceding BCs and interruptions only yields improvements for interruptions.

\subsection{Comparing Turn-Taking to SOTA Models}

In \cref{tab:turntaking}, we compare turn-taking behaviour to Moshi \citep{défossez2024moshispeechtextfoundationmodel} and dGSLM \citep{nguyen-etal-2023-generative}, with all metrics reported as cumulative durations per category. For all four metrics, Inter-Pausal Units (IPUs), intra-turn Pause, between-turn Gap and Overlap, our model exhibits behavior similar to the training data and the baseline models.  Like Moshi and non-cascaded dGLSM, our model’s pauses are generally longer than the gaps, reflecting real-world conversational statistics \citep{HELDNER2010555}. 

\begin{table}[t]
    \centering
    \resizebox{\linewidth}{!}{%
    \begin{tabular}{l l c c c c} 
        \toprule
        \textbf{Model} & & \textbf{IPU} & \textbf{Pause} & \textbf{Gap} & \textbf{Overl.} \\
        \midrule
        Best non-casc. & \multirow{3}{*}{\parbox{2cm}{\citet{nguyen-etal-2023-generative}}} & 41.4s&  13.8s & 10.7s &  6.1s\\
        Best casc. & & 54.8s &  0.0s&  5.3s&  0.0s \\
        Ground Truth & &53.5s&  5.5s&  4.4s & 3.6s \\
        \midrule
        Moshi & \multirow{2}{*}{\parbox{2cm}{\citet{défossez2024moshispeechtextfoundationmodel}}} & 50.8s &7.0s &  4.5s & 4.1s\\
        Ground Truth && 51.1s &  6.4s&  4.2s & 3.3s  \\
        \midrule
        Ours & (best)  & 59.3s & 10.4s & 3.0s & 5.4s \\
        Ground Truth &  (Beh.-SD)   & 55.8s & 10.8s &  3.8s & 3.0s \\
        \bottomrule
    \end{tabular}%
    }
    \caption{Cumulated durations per minute across models for turn-taking events.}
    \label{tab:turntaking}
\end{table}


\subsection{The Role of Sampling Parameters}\label{subsec:sampling_params}

Similar to \citet{défossez2024moshispeechtextfoundationmodel}, we observe that sampling temperature has a significant impact on model behaviour. Results for different temperatures ($0.6$–$1.0$) using our best-performing model (word-level alignment; audio delay of 2) on the dev set are reported in \cref{tab:sampling_param_results} in \cref{app:results}. As temperature increases, the dialogue becomes more dynamic, with more BCs and interruptions, more balanced speaking time between the speakers, and improved coherence and adherence to the narrative. However, higher temperatures also lead to slightly lower UTMOS scores and higher WER, reflecting a mild impact on speech quality. In our setup, a temperature of $0.9$ provides the best trade-off, and all results reported in this paper use this setting.

\subsection{RVQ vs FSQ}
Finally, we train our best-performing configuration (word-level alignment, audio delay of 2, s/u loss) using an RVQ encoder. Results are reported in \cref{tab:mimi_vs_nano} in \cref{app:results}. Although we do not explicitly model codebook dependencies and instead predict all codebooks in parallel, performance is comparable to that of the FSQ-based models. However, as expected, speech quality degrades substantially, with UTMOS dropping to 2.5 and similarly low scores observed for speaker embedding metrics.

\section{Conclusion}\label{sec:conclusion}
In this work, we introduce a framework for controllable full-duplex speech models that can be trained using less than 2,000 hours of speech. Despite the relatively limited training data and compute, our model \textit{F-Actor} produces coherent conversations with speech characteristics similar to the training data. It can be prompted with regards to conversation topic, speaker voice, and conversation initiation. Furthermore, we take steps toward making our model controllable with respect to backchannels and interruptions, allowing a user to decide how proactively the system should communicate with users. Finally, we systematically evaluate key design choices and release our model and code publicly, laying the foundation for future work on instruction-following full-duplex speech models.

\section{Limitations}\label{sec:limitations}
We identify the following main limitations of our work:
\begin{enumerate}
\item While the model’s predictions of backchannels and interruptions correlate with the numbers specified in the prompt, the model consistently produces fewer such events than requested. As discussed above, the prompt mainly serves as a directional signal indicating whether more or fewer backchannels or interruptions should occur, rather than enforcing an exact count. We believe this limitation is primarily due to the training data, in which backchannels and interruptions are relatively rare. Training on data with a higher density of these phenomena may allow the model to better learn their frequency and timing. Nonetheless, our results provide useful insights into how such behaviors can be modeled in full-duplex systems and point toward promising directions for future work.
\item The codec model we use, NanoCodec, does not currently support streaming encoder inference, which limits real-world deployment to chunk-wise processing rather than fully online generation.
\item We demonstrate the instruction-following capabilities of our full-duplex model in English conversations. While it would be very interesting to see whether our training recipe also works for other languages, this is currently infeasibly, as training data for other languages is not available \citep{chen2025turntakingsynchronousdialoguesurvey}.
\end{enumerate}

\section{Ethical Considerations}
Spoken conversational models pose inherent risks, including misuse for deception, impersonation, or social engineering. In particular, systems capable of natural, interactive speech could be exploited for scams or other forms of fraudulent behaviour.

We take several steps to mitigate these risks. First, the model is restricted to a small fixed pool of speaker embeddings, which prevents imitation of arbitrary or identifiable real-world voices. Second, the model is trained exclusively on text-to-speech (TTS) data, resulting in speech outputs that retain a synthetic character rather than fully natural human speech. This reduces the likelihood that the system could be mistaken for a real individual in high-stakes settings.

Finally, the model is released for research purposes only, with the goal of advancing work on controllable full-duplex conversational speech. We encourage responsible use and stress that safeguards and deployment policies are necessary for any real-world applications.

\section*{Acknowledgments}
This research is partially funded by the European Union’s Horizon research and innovation programme under grant agreement No. 101135798, project Meetween (My Personal AI Mediator for Virtual MEETtings BetWEEN People).
Ondrej Klejch was supported by the Scottish Government (Grant name: ``Ecosystem for Interactive Speech Technologies'') and a Turing AI Fellowship (Grant name: ``Neural Conversational Information Seeking Assistant'', EPSRC grant ref: EP/V025708/1). \\ 
We sincerely thank NVIDIA Corporation for their support via the ``NVIDIA Academic Grant Program''. 

\bibliography{custom}
\appendix
\crefalias{section}{appendix}
\crefalias{subsection}{appendix}
\crefalias{subsubsection}{appendix}
\section{Background: Speech and LLMs}\label{app:background}
To enable Large Language Models (LLMs) to process and generate audio, speech waveforms must be mapped into a discrete sequence of units. This compression serves two primary functions. First, it downsamples the high sample rate acoustic signal along the temporal axis to manageable sequence lengths. Second, it creates a finite vocabulary of units, allowing the model to be trained using standard cross-entropy objectives analogous to text-based language modeling. However, the properties of these units depend fundamentally on the learning paradigm used to derive them, broadly categorized into reconstruction-based or discriminative-based approaches.

\subsection{Discrete Representations of Speech}
The first category of discrete representations fo speech consists of units often termed ``acoustic tokens'' or Discrete Acoustic Units (DAUs), derived from reconstruction-based Neural Audio Codecs (NACs) such as Mimi, EnCodec, or SoundStream \citep{défossez2024moshispeechtextfoundationmodel, defossezhigh, zeghidour2021soundstream}. These units are optimized to compress and perfectly reconstruct the input signal. Consequently, they encode all acoustic information, including speaker identity, prosody, and recording conditions, into a single stream that can be decoded directly back into a waveform. While this enables end-to-end modeling such as in Moshi and VALL-E \citep{défossez2024moshispeechtextfoundationmodel, wang2023neural}, the high information density of DAUs can make modeling stability challenging.

In contrast, Discrete Speech Units (DSUs) are derived from discriminative Self-Supervised Learning (SSL) models like HuBERT or WavLM \citep{hubert_2021, chen2022wavlm}. Typically obtained via offline clustering of continuous hidden representations (e.g., via K-Means), DSUs correlate strongly with phonetic structure rather than low-level acoustics. While sometimes erroneously referred to as ``semantic tokens,'' they function effectively as a learned, phone-like vocabulary \citep{wells22_interspeech}. Unlike DAUs, DSUs tend to discard paralinguistic and prosodic information while preserving more phonetic content at lower bitrates. At higher bitrates, however, they allow for the restoration of paralinguistic and prosodic information at the cost of diminishing the phonetic interpretability of the discrete units \citep{sanders25_interspeech}. Although some previous work has decoded DSUs directly back into a waveform \citep{lakhotia-etal-2021-generative}, more recent work chooses to model DSUs with additional information like F0 \citep{kharitonov2022text,nguyen-etal-2025-spirit} or to use a two-stage or coarse-to-fine generation process. In this latter approach, a separate model generates continuous representations from DSUs before generating the waveform from those continuous representations \citep{anastassiou2024seed}.

\subsection{Quantization Methods} 
The choice of quantization method fundamentally dictates the architecture of the Speech Language Model, specifically regarding how it models the multiple codes representing a single time step. The two primary paradigms involve dependent codebooks, typically resulting from Residual Vector Quantization \citep[RVQ]{zeghidour2021soundstream}, and independent codebooks, such as those derived from Finite Scalar Quantization \citep[FSQ]{fsq_2024}. In RVQ-based models like EnCodec or Mimi \citep{defossezhigh, défossez2024moshispeechtextfoundationmodel}, the codebooks operate sequentially where each quantizes the residual error of the previous one. This creates a strong dependency among the codes within a single frame. Therefore, generation often requires a hierarchical approach. The primary language model autoregressively predicts the code for the first RVQ layer, while a secondary mechanism, such as the Depth Transformer in Moshi or the nonautoregressive model in VALL-E, predicts the codes for subsequent RVQ layers for that frame \citep{défossez2024moshispeechtextfoundationmodel, wang2023neural}. Conversely, methods like NanoCodec use FSQ to project the latent space into independent subspaces, resulting in codebooks that are statistically independent \citep{nanocodec_2025}. This independence simplifies the modeling task by allowing the Language Model to predict all codes for a frame simultaneously. Such models, as seen in \citet{hu25f_interspeech}, typically utilize a multiple head output layer, termed a Hydra head, which predicts the indices for all independent codebooks in parallel without the need for an additional sequential depthwise modeling step.

\section{Data}\label{app:data}

\subsection{Prompts}
To train the model as an instruction-following full-duplex system, we construct a prompt, similar to \citet{lee-etal-2025-behavior}, that specifies (i) the number of interruptions and backchannels in the conversation, provided by Behavior-SD; (ii) whether the system should initiate the dialogue or wait for the user; and (iii) the topic of the conversation. For the latter, we use the annotated narrative in Behavior-SD, e.g.,
\textit{Karina was playing her music loudly and Yoseph did not enjoy it. The bass was thumping and the lyrics were explicit. Yoseph felt annoyed and asked Karina to turn it down.}

We observed that in roughly 85\% of the cases the narrative begins with the name of the speaker who also starts the conversation. In early experiments, this caused the model to ignore the explicit instruction about who should speak first, as it relied instead on the narrative. To prevent this, we rewrite the narrative to be system-centric. In doing so, we also replace the system’s name with \textit{you}, which encourages the model to interpret the narrative from the system’s own perspective. For instance, if the system corresponds to \textit{Karina}, the rewritten version becomes:
\textit{You were playing your music loudly and Yoseph did not enjoy it. The bass was thumping and the lyrics were explicit. Yoseph felt annoyed and asked you to turn it down.}
We use \texttt{google/gemma-1.1-7b-it}\footnote{\url{https://huggingface.co/google/gemma-1.1-7b-it}} \citep{gemmateam2024gemmaopenmodelsbased} to perform this rewriting on the train set and GPT-5.1 on the test set.

As an alternative formulation, we also experiment with goal-oriented prompting, in which the system does not receive a narrative but instead a set of goals it should accomplish during the dialogue. These goal descriptions are likewise generated using the same LLM.

Example prompts can be found in \cref{fig:app_prompts}.
\begin{figure}[ht]
\textbf{System Prompt using Narrative:}
\small{\begin{lstlisting}[breaklines=true, breakindent=0pt]
Generate a dialogue between you (Karina) and another speaker (Yoseph) based on the given narrative. Follow the specific behavior instructions for you.

Narrative:
- You were playing your music loudly and Yoseph did not enjoy it. The bass was thumping and the lyrics were explicit. Yoseph felt annoyed and asked you to turn it down.

Your behaviors:
- backchannels: 9
- interruptions: 0
- starts the dialogue: False

Ensure that the dialogue reflects the behaviours of you.
\end{lstlisting}}
    \caption{Example prompts from the train set.}
    \label{fig:app_prompts}
\end{figure}





\subsection{Rewriting Narratives}
Rewriting prompts can be found in \cref{fig:app_rewriting_prompts}.
\begin{figure*}[ht]
\textbf{Prompt to rewrite narrative to system/user style.}
\small{\begin{lstlisting}[breaklines=true, breakindent=0pt, basicstyle=\ttfamily]
Your task is to rewrite a written narrative from the perspective of a specified speaker, 
replacing only that speaker's name and pronouns with 'you' and 'your'. 
Do NOT change other characters or their pronouns. 
Do NOT change the narrative events or add explanations. Maintain the narrative flow and adjust active/passive voice as needed.

Example:
Input: Replace Karina and corresponding pronouns with you and your. 
Karina was playing her music loudly and Yoseph did not enjoy it. 
The bass was thumping and the lyrics were explicit. Yoseph felt annoyed and asked Karina to turn it down.
Output: You were playing your music loudly and Yoseph did not enjoy it. 
The bass was thumping and the lyrics were explicit. Yoseph felt annoyed and asked you to turn it down.

New Input Narrative:
Replace {speaker} and corresponding pronouns with you and your.
{narrative}
Output Narrative:
\end{lstlisting}}
    \caption{Prompt to rewrite the narrative from Behavior-SD.}
    \label{fig:app_rewriting_prompts}
\end{figure*}




\section{Evaluation}\label{app:eval}

\subsection{General System Capabilities Eval.}
To assess general system capabilities, we report the perplexity on both the speech and text streams. Speech quality is evaluated using UTMOS \citep{utmos_2022}, a model that predicts mean opinion scores and has been applied previously to other full-duplex models \citep{arora2025chainofthoughtreasoningstreamingfullduplex, zhang-etal-2025-omniflatten}. We first use Parakeet\footnote{\huggingfacesmall{} \href{https://huggingface.co/nvidia/parakeet-tdt-0.6b-v2}{nvidia/parakeet-tdt-0.6b-v2}} to obtain segment-level timestamps for each speaker, calculate UTMOS for each segment, and then average these scores across segments and speakers to obtain the final dialogue-level score. Additionally, we measure the speaking-time difference between the two speakers to ensure balanced participation and that no single speaker dominates the conversation. Finally, we transcribe the generated audio using Parakeet and calculate the word-error rate (WER) against the generated text stream to evaluate the alignment between the model’s speech and text outputs. We remove special tokens, such as dedicated backchannel and interruption tokens before calculating WER.

\paragraph{Instruction-Following Capabilities Evaluation}
We evaluate the model’s instruction-following along four dimensions:

\begin{enumerate}
    \item \textbf{Speaker Initiation}: The ratio of the correct speaker starting the conversation according to the prompt. We determine the first speaker using Parakeet-generated segment \footnote{\huggingfacesmall{} \href{https://huggingface.co/nvidia/parakeet-tdt-0.6b-v2}{nvidia/parakeet-tdt-0.6b-v2}}.
    \item \textbf{Speaker Embedding Consistency:} Cosine similarity between the target speaker embedding and the generated speech for each speaker, averaged across speakers. To compute this, we randomly sample snippets of 3–5 seconds from each dialogue, encode them with ECAPA-TDNN \citep{dawalatabad2021ecapa} , and compare to the target embeddings. To assess potential speaker drift over the dialogue, we also calculate the distance (1–cosine similarity) between the first and last segments for each speaker and average across speakers.
    \item 
    \textbf{Narrative Adherence:} An LLM judge (\texttt{Llama-3.1-8B-Instruct}\footnote{\huggingfacesmall{} \href{https://huggingface.co/meta-llama/Llama-3.1-8B-Instruct}{meta-llama/Llama-3.1-8B-Instruct}} \citep{grattafiori2024llama3herdmodels}
    ) evaluates the alignment (of the Parakeet transcript) with the narrative specified in the prompt. Prompts for the LLM judge are provided in \cref{fig:app_llm_judge}.
    \item \textbf{Backchannels and Interruptions:} We measure the number of backchannels and interruptions per speaker and report correlations with the prompt-specified counts. Pearson’s correlation coefficient (r) quantifies the linear relationship, and two-sided p-values are computed using the exact distribution for the null hypothesis $r>0$.  More details on the detection of backchannels and interruptions can be found in \cref{app:bc_expl}.
\end{enumerate}

\subsection{Counting Backchannels and Interruptions}\label{app:bc_expl}

For evaluating backchannels and interruptions, we experiment with several detection algorithms. We initially adopt the FD-Bench \citep{peng25b_interspeech} implementation, which uses Silero-VAD \citep{Silero_VAD} to obtain speech timestamps. However, default thresholds for merging words into utterances and defining backchannels or interruptions did not generalize well to the Behavior-SD test set, often misclassifying interruptions or assigning incorrect timestamps to short backchannels. Detecting these events is challenging due to their short duration, so sometimes they are not transcribed or assigned a timestamp at all.

To address this, we compare timestamps obtained from Silero-VAD \citep{Silero_VAD}, Parakeet\footnote{\huggingfacesmall{} \href{https://huggingface.co/nvidia/parakeet-tdt-0.6b-v2}{nvidia/parakeet-tdt-0.6b-v2}}, and Vosk\footnote{\vosksmall \href{https://alphacephei.com/vosk/models/vosk-model-en-us-0.22.zip}{vosk-model-en-us-0.22}}. We perform a grid search over three parameters:

\begin{enumerate}
    \item Split threshold: Controls when consecutive words are merged into a single utterance, varied from 0.20 to 0.90 in steps of 0.005.
    \item Interruption threshold: Specifies how close to the end of a conversation a segment must occur to be considered an interruption, varied from 0.10 to 0.70 in steps of 0.005.
    \item Overlap tolerance: Defines the maximum allowed temporal overlap between segments of the same speaker for them to be treated as overlapping speech rather than separate utterances, varied from 0.05 to 0.50 in steps of 0.005.
\end{enumerate}
The best-performing configuration uses a split threshold of $0.565$, an interruption threshold of $0.405$, and an overlap tolerance of $0.435$, which we adopt for all subsequent evaluations.

\begin{table}[!ht]
    \centering
    \resizebox{\linewidth}{!}{%
    \begin{tabular}{llccccc}
        \toprule
        \multirow{2}{*}{\textbf{Model}} & 
        \multirow{2}{*}{\textbf{Segm.}} &
        \multicolumn{2}{c}{\textbf{Missing} $\downarrow$} &
        \multicolumn{2}{c}{\textbf{Extra} $\downarrow$} \\
        \cmidrule(lr){3-4} \cmidrule(lr){5-6}
        & & \textbf{BC} & \textbf{Inter.} &\textbf{BC} & \textbf{Inter.} \\
        \midrule
        \multirow{2}{*}{Parakeet} & word & \textbf{1.0} $\pm$ \textbf{1.4} & \textbf{0.5} $\pm$ \textbf{0.7} &\textbf{0.2} $\pm$  0.4 & 0.5 $\pm$ 0.7 \\
         & segm. & 2.7 $\pm$ 2.9 & 1.0 $\pm$ 1.0 & 3.0 $\pm$ 2.9 & 1.0 $\pm$ 1.9\\
        \multirow{2}{*}{Kaldi} & word & 1.7 $\pm$ 1.9 & 0.8 $\pm$ 0.9 & 1.1 $\pm$ 1.6 & \textbf{0.4} $\pm$ \textbf{0.6} \\
        
        & segm. & 1.3 $\pm$ 1.5 & 0.9 $\pm$ 0.9 & 0.7 $\pm$ 0.7 & 0.6 $\pm$ 0.7\\
        Silero-VAD & segm. &  1.6 $\pm$ 1.7 & 1.0 $\pm$ 1.0 & 0.3 $\pm$ 0.6 & 0.8 $\pm$ 1.0 \\

        \bottomrule
    \end{tabular}%
    }
    \caption{Results of interruption (\textit{Inter.}) and backchannel (\textit{BC}) detection algorithms on the Behavior-SD test set, averaged per dialogue across the test set. For each model, we report performance using its native segmentation (\textit{segm.}) and using word-level timestamps merged with our grid-searched parameters (\textit{word}). Silero-VAD uses the default FD-Bench settings \citep{peng25b_interspeech}. 
    }
    \label{tab:bc_inter_algo}
\end{table}

Parakeet provides the most reliable performance on Behavior-SD \citep{lee-etal-2025-behavior}, missing on average only 1.0 backchannels and 0.5 interruptions per conversation. This means, $79.4$\% of the backchannels are correctly classified, and $79.2$\% of the interruptions are correctly classified.
Results for all alignment methods using these threshold are reported in \cref{tab:bc_inter_algo}. Based on these findings, we adopt Parakeet for all subsequent evaluations.

\begin{figure*}[ht]
\textbf{System Prompt for LLM Judge:}
\small{\begin{lstlisting}[breaklines=true, breakindent=0pt, basicstyle=\ttfamily]
You are an expert evaluator of dialogues. 
You only respond with a single digit from 1 to 5 based on the evaluation criteria.
\end{lstlisting}}
\normalsize\textbf{User Prompt to Judge Dialogue Coherence:}
\small{\begin{lstlisting}[breaklines=true, breakindent=0pt, basicstyle=\ttfamily]
Evaluate how well the following dialogue fits the given narrative.

Criteria:
1. Relevance: Does the dialogue clearly reflect the situation or topic described in the narrative?
2. Consistency: Are the characters, events, and tone in the dialogue consistent with the narrative?
3. Faithfulness: Does the dialogue avoid introducing contradictions or unrelated content?
- Do NOT judge fluency or engagement - only topical/narrative alignment.
- Score strictly between 1 and 5 (1 = Not related at all, 5 = Perfectly fits the narrative).

Narrative:
{narrative}

Dialogue:
{transcription}

Score:
\end{lstlisting}}

\caption{Prompt for the LLM Judge to judge the instruction following capabilities of our full-duplex model.}
\label{fig:app_llm_judge}
\end{figure*}

\subsection{Human Evaluation to Confirm the LLM Judge Findings}\label{app:human_eval}
To confirm the reliability of the LLM judge, we conduct a human evaluation. 

\parag{Setup.} Annotators were given the same rating criteria as the LLM judge, evaluating each dialogue on its conversational coherence and faithfulness to the given narrative. We selected our two best-performing models (both using s/u loss, speaker embeddings, word alignment, and an audio delay of 2; one also incorporating BC and interruption tokens) alongside the Behaviour-SD original dialogues. Using the Pearmut platform \citep{zouhar2026pearmuthumanevaluationtranslation}, 12 annotators each rated 60 samples (20 dialogues × 3 models), with every dialogue annotated by 3 annotators, yielding 240 annotated instances across 80 unique dialogues.

\parag{Annotators.} Annotators were 12 computer science researchers (5 female, 7 male), compensated according to the national collective wage agreement. Participants were informed that their ratings would be published, while their identities remain anonymous. No GDPR-protected personal data was collected. The annotators geographic location was in Europe.

\parag{Duration.} The average annotation session (20 dialogues × 3 models) took 56.1 minutes per annotator, underscoring the practical necessity of automatic evaluation metrics for these kinds of evaluations.

\begin{figure*}
    \centering
    \includegraphics[width=1.0\linewidth]{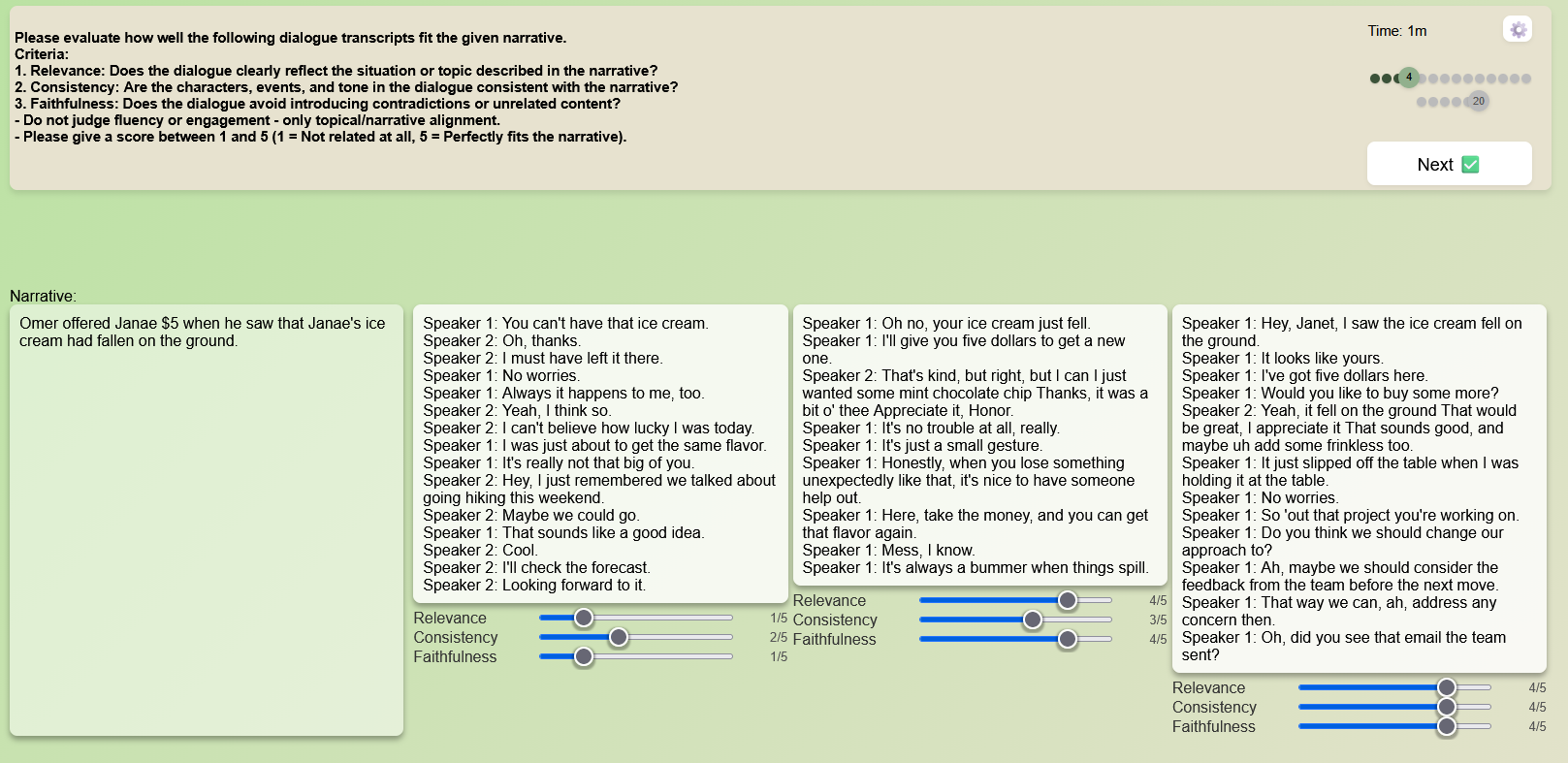}
    \caption{Screenshot of the instructions and interface for the human evaluation.}
    \label{fig:human_eval}
\end{figure*}

The full annotation instructions and a platform screenshots can be found in \cref{fig:human_eval}.
The files used to set up the evaluation are also available in the github repository of this project.

\section{Training Details}\label{app:training}
Training and inference parameters are shown in \cref{tab:hyperparams}. We train our models on four NVIDIA A100-SXM4-40GB. Training takes around 46-48 hours, depending on the configuration. 

For inference, we show that sampling parameters, in particular the temperature, have a great influence on model performance in \cref{subsec:sampling_params}. We find that for our setup, a temperature of $0.9$ works best and hence report all results (except for those in \cref{tab:sampling_param_results}) with this temperature.
\begin{table}[!ht]
    \centering
    \begin{tabular}{lc}
    \toprule
    \textbf{Hyperparameter} & \textbf{Value} \\
    \midrule
    Train Batch Size (micro-batch) & 1 \\
    Gradient Accumulation Steps & 8 \\
    Learning Rate & 5e-5 \\
    Max Steps (optimizer updates) & 100{,}000 \\
    Examples per Step & 32 \\
    Max Sequence Length & 2048 \\
    Precision & bfloat16 \\
    Audio Vocab Size & 4032 \\
    Early Stopping Patience & 10 \\
    Gradient Clipping & 1.0 \\
    Weight Decay & 0.01 \\
    \midrule
    Max Sequence Length & 1024  \\
    Temperature & 0.9* \\
    Top-k & 40 \\
    Top-p & 1 \\
    \bottomrule
    \end{tabular}
    \caption{Training and inference hyperparameters for training our full-duplex model using DeepSpeed. We use four NVIDIA A100-SXM4-40GB GPUs for training.\\
    *Unless explicitly stated otherwise.}
    \label{tab:hyperparams}
\end{table}

\begin{table*}[ht]
    \centering
    \resizebox{\linewidth}{!}{%
    \begin{tabular}{ccccccccccccc}
        \toprule
        \multirow{2}{*}{\textbf{Temperature}} &
        \textbf{UTMOS} &
        \textbf{WER \%} &
        \textbf{Avg. Speaking} &
        \textbf{Correct Start} & \textbf{Spk. Sim.} & \textbf{Spk. Drift} &
         \textbf{Narrative} &
        \textbf{BC Corr.*} & \textbf{Inter. Corr.*} \\
        &
        $\uparrow$ &
        \textbf{audio/text $\downarrow$} &
        \textbf{Diff (s) $\downarrow$} 
        &
        (\% $\uparrow$) & (cos $\uparrow$) & (1-cos $\downarrow$) &
        (1--5 $\uparrow$)  &
        (per dial. $\uparrow$) & (per dial.$\uparrow$) \\
         \midrule
          \multicolumn{1}{c}{Behaviour-SD} & 3.78 &  4.5 & 15.44 & 100 & 0.62 & 0.62 & 4.04 & 0.92 & 0.74 \\
         \multicolumn{1}{c}{Lower Baseline} & -& -& - & 50$^a$ & 0.35$^b$ & 0.75$^c$ &  1.26$^d$ & - & -\\
         \midrule
0.6 & \cellcolor[HTML]{a4c2f4}\textcolor[HTML]{000000}{3.48} & \cellcolor[HTML]{a4c2f4}\textcolor[HTML]{000000}{6.2} & \cellcolor[HTML]{ffffff}\textcolor[HTML]{000000}{24.29} & \cellcolor[HTML]{a4c2f4}\textcolor[HTML]{000000}{100.0} & \cellcolor[HTML]{a4c2f4}\textcolor[HTML]{000000}{0.54} & \cellcolor[HTML]{a4c2f4}\textcolor[HTML]{000000}{0.59} & \cellcolor[HTML]{bfd4f7}\textcolor[HTML]{000000}{2.57} & \cellcolor[HTML]{ffffff}\textcolor[HTML]{000000}{0.30} & \cellcolor[HTML]{ffffff}\textcolor[HTML]{000000}{0.13} \\
0.8 & \cellcolor[HTML]{c8daf8}\textcolor[HTML]{000000}{3.44} & \cellcolor[HTML]{c2d6f8}\textcolor[HTML]{000000}{7.5} & \cellcolor[HTML]{c2d6f8}\textcolor[HTML]{000000}{16.12} & \cellcolor[HTML]{a4c2f4}\textcolor[HTML]{000000}{100.0} & \cellcolor[HTML]{a4c2f4}\textcolor[HTML]{000000}{0.54} & \cellcolor[HTML]{ffffff}\textcolor[HTML]{000000}{0.63} & \cellcolor[HTML]{e8effc}\textcolor[HTML]{000000}{2.26} & \cellcolor[HTML]{bed4f7}\textcolor[HTML]{000000}{0.47} & \cellcolor[HTML]{bad1f7}\textcolor[HTML]{000000}{0.22} \\
0.9 & \cellcolor[HTML]{edf3fd}\textcolor[HTML]{000000}{3.40} & \cellcolor[HTML]{d6e4fa}\textcolor[HTML]{000000}{8.35} & \cellcolor[HTML]{b2cbf6}\textcolor[HTML]{000000}{13.93} & \cellcolor[HTML]{c2d6f8}\textcolor[HTML]{000000}{99.8} & \cellcolor[HTML]{a4c2f4}\textcolor[HTML]{000000}{0.54} & \cellcolor[HTML]{d1e0f9}\textcolor[HTML]{000000}{0.61} & \cellcolor[HTML]{a4c2f4}\textcolor[HTML]{000000}{2.78} & \cellcolor[HTML]{a4c2f4}\textcolor[HTML]{000000}{0.54} & \cellcolor[HTML]{a4c2f4}\textcolor[HTML]{000000}{0.25} \\
1.0 & \cellcolor[HTML]{ffffff}\textcolor[HTML]{000000}{3.38} & \cellcolor[HTML]{ffffff}\textcolor[HTML]{000000}{10.09} & \cellcolor[HTML]{a4c2f4}\textcolor[HTML]{000000}{12.07} & \cellcolor[HTML]{ffffff}\textcolor[HTML]{000000}{99.4} & \cellcolor[HTML]{ffffff}\textcolor[HTML]{000000}{0.53} & \cellcolor[HTML]{ffffff}\textcolor[HTML]{000000}{0.63} & \cellcolor[HTML]{ffffff}\textcolor[HTML]{000000}{2.08} & \cellcolor[HTML]{abc7f5}\textcolor[HTML]{000000}{0.52} & \cellcolor[HTML]{bad1f7}\textcolor[HTML]{000000}{0.22} \\
         \bottomrule
         \multicolumn{11}{l}{
        $^{a}$ random spk. starts; 
        $^{b}$ swapped spk. embeddings; 
        $^{c}$ first embedding = spk. 1, last = spk. 2; 
        $^{d}$ random narrative.} \\
         \multicolumn{11}{l}{
        $*$ All correlations are statistically significant with $p<0.01$}\\
    \end{tabular}%
    }
    \caption{Instruction-following results for our best model (text-stream, audio delay of 2, word-alignment, trained on system + user) for different sampling parameters.}
    \label{tab:sampling_param_results}
\end{table*}
\begin{table*}[ht]
    \centering
    \resizebox{\linewidth}{!}{%
    \begin{tabular}{ccccccccccccc}
        \toprule
        \multirow{2}{*}{\textbf{FSQ/RVQ}} &
        \textbf{UTMOS} &
        \textbf{WER \%} &
        \textbf{Avg. Speaking} &
        \textbf{Correct Start} & \textbf{Spk. Sim.} & \textbf{Spk. Drift} &
         \textbf{Narrative} &
        \textbf{BC Corr.*} & \textbf{Inter. Corr.*} \\
        &
        $\uparrow$ &
        \textbf{audio/text $\downarrow$} &
        \textbf{Diff (s) $\downarrow$} 
        &
        (\% $\uparrow$) & (cos $\uparrow$) & (1-cos $\downarrow$) &
        (1--5 $\uparrow$)  &
        (per dial. $\uparrow$) & (per dial.$\uparrow$) \\
         \midrule
          \multicolumn{1}{c}{Behaviour-SD} & 3.78 &  4.5 & 15.44 & 100 & 0.62 & 0.62 & 4.04 & 0.92 & 0.74 \\
         \multicolumn{1}{c}{Lower Baseline} & -& -& - & 50$^a$ & 0.35$^b$ & 0.75$^c$ &  1.26$^d$ & - & -\\
         \midrule
        FSQ (NanoC.) & \textbf{3.40} &  8.35&  13.93 &  \phantom{0}99.8 & \textbf{0.54} & 0.61  &2.78&  \textbf{0.54} & \textbf{0.25}\\
        RVQ (MimiC.) & 2.50 & \textbf{7.18} & \textbf{11.12} & \textbf{100.0} & 0.40 & \textbf{0.59} & \textbf{2.83} & 0.40 & 0.25\\
         \bottomrule
         \multicolumn{11}{l}{
        $^{a}$ random spk. starts; 
        $^{b}$ swapped spk. embeddings; 
        $^{c}$ first embedding = spk. 1, last = spk. 2; 
        $^{d}$ random narrative.} \\
         \multicolumn{11}{l}{
        $*$ All correlations are statistically significant with $p<0.01$}\\
    \end{tabular}%
    }
    \caption{Instruction-following results for our best model (text-stream, audio delay of 2, word-alignment, trained on system + user) trained with FSQ or RVQ speech codecs.}
    \label{tab:mimi_vs_nano}
\end{table*}

\section{Results}\label{app:results}
We report and discuss results for all of our models in \cref{sec:results}, with two exceptions, that are reported in the appendix:
Detailed results using different temperatures during inference are shown in \cref{tab:sampling_param_results}.
Results comparing models using NanoCodecs (FSQ) and Mimi (RVQ) audio encoding are shown in \cref{tab:mimi_vs_nano}.

\end{document}